\documentclass[10pt,twocolumn,letterpaper]{article}

\usepackage{cvpr}              % To produce the CAMERA-READY version
\usepackage{graphicx}
\usepackage{amsmath}
\usepackage{amssymb}
\usepackage{booktabs}

\usepackage{comment}
\usepackage{bbm}
\usepackage{amsmath}
\usepackage[pagebackref,breaklinks,colorlinks]{hyperref}

\usepackage[capitalize]{cleveref}
\crefname{section}{Sec.}{Secs.}
\Crefname{section}{Section}{Sections}
\Crefname{table}{Table}{Tables}
\crefname{table}{Tab.}{Tabs.}

\def\cvprPaperID{*****} % *** Enter the CVPR Paper ID here
\def\confName{CVPR}
\def\confYear{2022}

\begin{document}

%%%%%%%%% TITLE - PLEASE UPDATE
\title{Deep Level Set for Box-supervised Instance Segmentation in Aerial Images}

%\title{Deep Level Set for Box-supervised Instance Segmentation \\ in Aerial Images}

\author{Wentong Li$^1$, Yijie Chen$^1$, Wenyu Liu$^1$, Jianke Zhu$^{1,2}\thanks{Corresponding author is Jianke Zhu.}$\\
	$^1$Zhejiang University \\
	$^2$Alibaba-Zhejiang University Joint Research Institute of Frontier Technologies \\ 
	{\tt\small \{liwentong, chen\_yj, liuwenyu.lwy, jkzhu\}@zju.edu.cn}}
\maketitle

%%%%%%%%% ABSTRACT
\begin{abstract}
   Box-supervised instance segmentation has recently attracted lots of research efforts while little attention is received in aerial image domain. In contrast to the general object collections, aerial objects have large intra-class variances and inter-class similarity with complex background. Moreover, there are many tiny objects in the high-resolution satellite images. This makes the recent pairwise affinity modeling method inevitably to involve the noisy supervision with the inferior results. To tackle these problems, we propose a novel aerial instance segmentation approach, which drives the network to learn a series of level set functions for the aerial objects with only box annotations in an end-to-end fashion. Instead of learning the pairwise affinity, the level set method with the carefully designed energy functions treats the object segmentation as curve evolution, which is able to accurately recover the object's boundaries and prevent the interference from the indistinguishable background and similar objects. The experimental results demonstrate that the proposed approach outperforms the state-of-the-art box-supervised instance segmentation methods. The source code is available at \href{https://github.com/LiWentomng/boxlevelset}{https://github.com/LiWentomng/boxlevelset.}
\end{abstract}

%%%%%%%%% BODY TEXT
\section{Introduction}
\label{sec:intro}

An insightful understanding of the objects (e.g. buildings, vehicles, planes, ships etc) from the remote sensing images with high spatial resolution (HSR)  has great value in many real-world applications inluding environmental monitoring and urban management~\cite{cvpr2018dota}. 

\begin{figure}[t]
	\centering
	\includegraphics[width=1.0\linewidth]{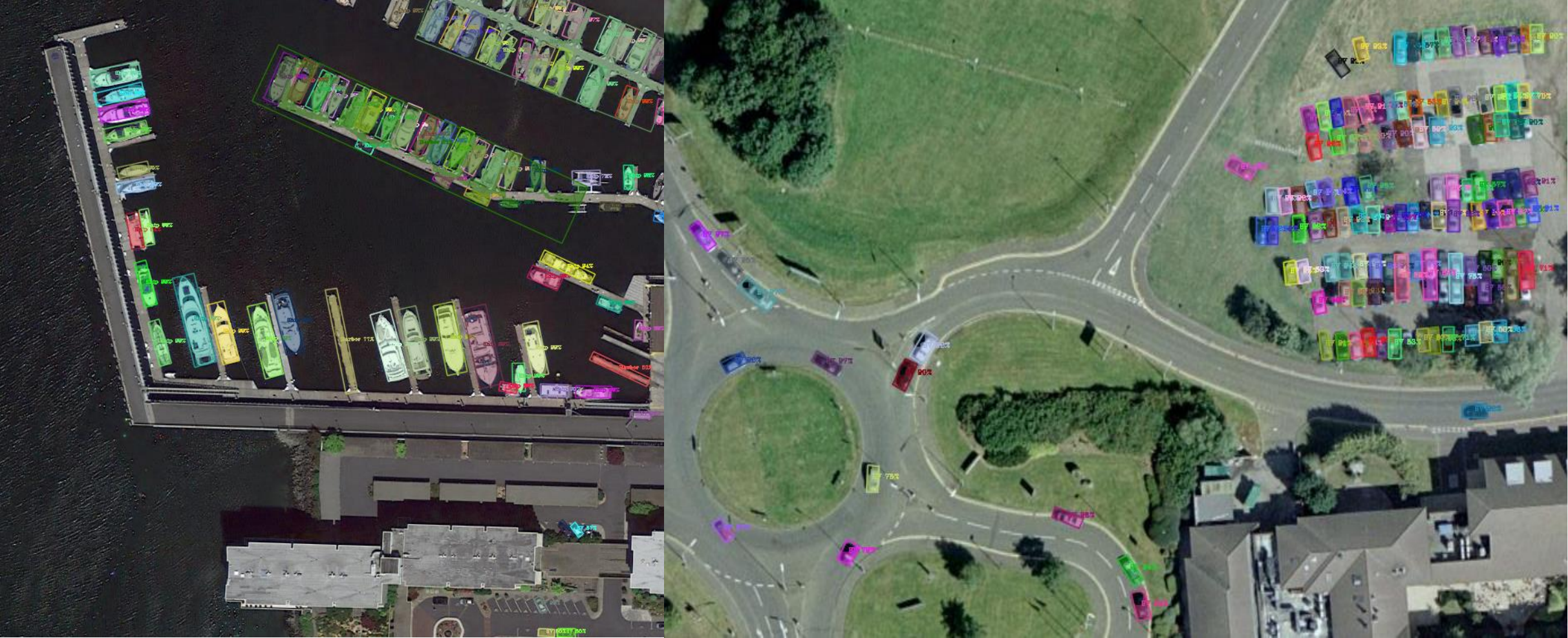}
	\caption{Illustration of intra-class variance, inter-class similarity and many tiny objects with complex background in aerial images.}\label{intro}
\end{figure}

Aerial instance segmentation is an important task having the pixel-level object localization rather than the bounding box, which can provide the detailed shape and pose cues for the interest objects~\cite{cvpr2020foreground}. Differently from general object collections~\cite{eccv2014coco,  ijcv2010_voc}, aerial images have their own difficulties including the larger intra-class variance and inter-class similarity with complex background~\cite{cvpr2021pointflow}, many tiny objects in the high resolution satellite images~\cite{cvpr2020foreground}. Figure~\ref{intro} shows two visual examples of aerial instance segmentation results.  

Although having achieved the promising results, most of previous instance segmentation methods heavily depend on the pixel-wise mask annotations that is notoriously time-consuming comparing to box-level annotations~\cite{eccv2016_pointsupervision, eccv2020box2seg}. Instead of relying on the pixel-level labels, box-supervised instance segmentation has recently attracted lots of research efforts~\cite{nips2019-bbtp, cvpr2020_boxinst}, which are designed with the pairwise affinity modeling, e.g. neighboring pixel pairs~\cite{nips2019-bbtp}, color pairs~\cite{cvpr2020_boxinst} etc. However, the pairwise affinity-based methods are defined on the set containing all or partial neighboring pixel pairs, which oversimplifies the assumption of spatially pixel or color pairs being encouraged to share the same label. This inevitably introduces the heavy noisy supervision in aerial instance segmentation. BoxInst~\cite{cvpr2020_boxinst} models the pairwise color affinity on an undirected graph built from the input image. BBTP~\cite{nips2019-bbtp} makes use of the pairwise pixel and formulates the box-supervised instance segmentation into a multiple instance learning problem. The above pairwise affinity learning may inevitably absorb the noisy context from the nearby background or similar objects, which leads to the inferior instance segmentation performance in aerial images.

In this paper, we propose a novel approach to aerial instance segmentation using box annotations, which is based on level set~\cite{osher1988fronts, tip2001_active_contour}. Unlike the simple pairwise affinity modeling, our method implicitly evolves a closed curve to directly recover the object boundaries, which is able to effectively reduce the noisy interference. In contrast to the previous methods managing to evolve the level set to the ground truth boundary in a fully-supervised manner~\cite{CVPR2017hudeep,cvpr2019deeplevelsetevolution, eccv2020levelset}, our proposed approach presents an end-to-end network weakly supervised by bounding box annotations originally for detection. It consists of detection branch and segmentation branch, which adopt a unified way to select the potential positive samples, i.e. box samples for detection branch and mask map samples for segmentation branch. 
We train classification, box regression and mask-level prediction jointly. In addition, the whole evolution process of level-set function is differentiable. Within the enlarged ground-truth box region, the level sets are iteratively updated to optimize the our presented energy function, which forces the boundary to evolve so as to accurately segment object from the background during the training. To enable the efficient convergence of evolution process, we introduce an effective box and background constraint.

In summary, the main contributions of this work are: 1) a novel deep level set-based approach to aerial instance segmentation. To the best of our knowledge, this is the first deep level set-based method that tackles the problem of box-supervised instance segmentation; 2) a framework to learn the potential positive samples, i.e.  boxes and mask map samples, in a unified manner, which can be jointly trained to achieve classification, box regression and mask-level prediction; 3) the promising aerial instance segmentation results on the benchmark using box annotations, which is comparable with the fully mask-supervised methods in some categories.

\section{Related Work}
\subsection{Level Set-based Segmentation}
The level set method~\cite{levelset1995a, osher1988fronts} is widely used in image segmentation due to its ability of automatically finding the object's boundaries. The key idea is to define an energy function in a higher dimension to represent the implicit curve, which is iteratively evolved by the descent of gradients. Later, Chan~\textit{et al.}~\cite{tip2001_active_contour} introduce a region-based level set method having the energy function with the curve inside and outside, which is more robust to the complex background or initial contour.

Deep network is capable of effectively encoding the high-level features. Therefore, some works~\cite{eccv2020levelset,tip2019mumford, CVPR2017hudeep,cvpr2019deeplevelsetevolution, WACV2019_semantideeplevelset} embed the level set into deep network, which achieve the promising segmentation results. For the semantic segmentation task, Le~\textit{et al.}~\cite{tip2018reformulating} propose the Contextual Recurrent Level Set (CRLS) method, which is presented in a time series for the curve evolution and is reformulated as Recurrent Neural Network. Kim~\textit{et al.} ~\cite{WACV2019_semantideeplevelset} convert level set functions into the maps of class probability and calculate the energy for each class, which can obtain the multi-class segmentation in an image. For instance annotation, Wang~\textit{et al.}~\cite{cvpr2019deeplevelsetevolution} propose a method that predicts the evolution parameters and evolves the predicted initial contour by incorporating the user clicks on the extreme boundary points. Levelset R-CNN~\cite{eccv2020levelset} embeds the Chan-Vese levelset segmentation optimization on top of Mask R-CNN~\cite{iccv2017maskrcnn} to facilitate the accurate instance segmentation. In spite of the promising performance, these methods adopt the level set to help deep network accurately evolve to the ground-truth boundary in a fully supervised manner. This paper exploit the deep level set on the task of box-supervised instance segmentation. ~\cite{tip2019mumford} is the most related work to ours, which focuses on the global pixel similarity between the Mumford–Shah function~\cite{mumford1989optimal} with multi-level sets and N-class softmax characteristic to achieve image segmentation in a semi-supervised or unsupervised manner. In this work, our method makes use of the curve evolution within its local box region for each object. The energy function is different and the levelset function in our work evolves the foreground object specially on single potential mask map during the optimization.

\begin{figure*}[t]
	\centering
	\includegraphics[width=0.95\linewidth]{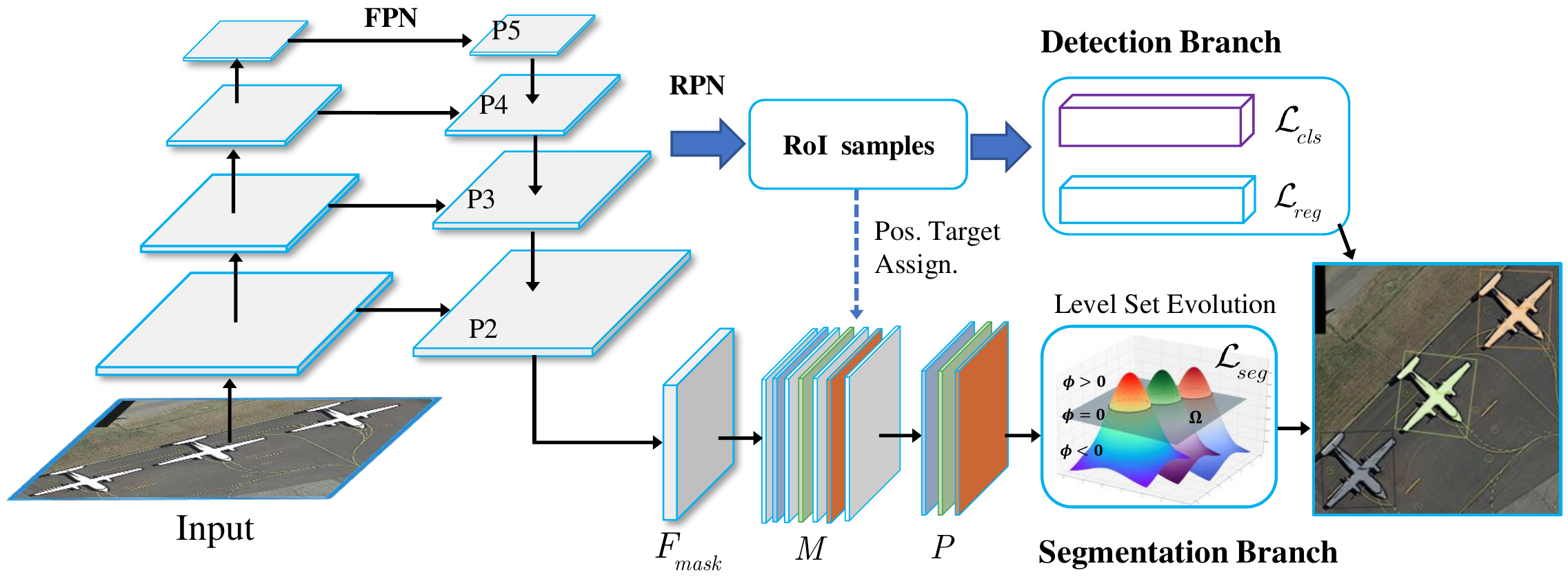}
	\caption{The overall architecture of the proposed method. The backbone with FPN is regarded as feature encoder of the input image. With the detection branch and segmentation branch, the accurate detection and instance segmentation are obtained through the bounding box annotations only. }\label{fig1overallnetwork}
\end{figure*}

\subsection{Box-Supervised Instance Segmentation}
Modern instance segmentation methods~\cite{iccv2017maskrcnn, cvpr2021refinemask, iccv2019yolact, eccv202_condinst, wang2020solov2} with fully mask-level supervision are able to segment objects with the accurate boundaries,  
%can be divided into two categories. The first group utilizes a powerful detector to generate the bounding boxes, and then performs instance segmentation on the detected regions~\cite{iccv2017maskrcnn,cvpr2020pointrend, cvpr2021refinemask}. Another category is to directly segment each instance without resorting to the detection results~\cite{iccv2019yolact, eccv202_condinst}. 
%Although being able to segment objects with the accurate boundaries,
while these methods rely on the pixel-wise mask annotations. This may limit their deployments in the real-world applications.
Box-supervision instance segmentation methods recently attract more attention in the general scene. SDI~\cite{cvpr2017SDI} is 
the first method to predict the mask with the box annotations under the deep learning framework. It is heavily dependent on the region proposals generated by the unsupervised segmentation methods like GrabCut~\cite{TOG2004grabcut} and MCG~\cite{tpmai2017mcg}. %Then an iterative training procedure are performed to gain a better segementation result.
Most recently, BBTP~\cite{nips2019-bbtp} models the neighboring pixel-pairwise affinity and defines the box-supervised instance segmentation as the multiple instance learning (MIL) based on the Mask R-CNN~\cite{iccv2017maskrcnn}.
BoxInst~\cite{cvpr2020_boxinst} uses the color-pairwise affinity modeling with box projection, which is built on an efficient RoI-free CondInst framework~\cite{eccv202_condinst}. Although having achieved the promising performance in the general scene, their pairwise affinity modeling is defined on either the set containing  partial or all neighboring pixel pairs with the oversimplified assumption of spatially pixel or color pairs being encouraged to share the same label. This inevitably introduces the heavily noisy supervision in the aerial image scene. Besides, the recently advanced methods BBAM~\cite{cvpr2021bbam} and BoxCaseg~\cite{cvpr2021boxcaseg} introduce multi training and inference stages or additional supervised information such as the mask guideline from the salient images to achieve promising segmentation performance. Differently from the above methods, our proposed level set-based method evolves implicitly in an end-to-end manner, which can directly learn to align with the instance's boundaries and prevent the noisy interference effectively within its box region. 

\subsection{Instance Segmentation in Aerial Image}
In contrast to the general scene, high-resolution remote sensing image has its unique characteristics, such as intra-class variance, large inter-class similarity with complex background~\cite{cvpr2021pointflow}, many tiny objects in high resolution images~\cite{cvpr2020foreground}, and large object-scale variations~\cite{cvprw2019isaid, igars2019multi-yolt}. Therefore, it is difficult for the conventional instance segmentation methods to directly deal with the remote sensing imagery. Mou~\textit{et al.}~\cite{tgrs2018vehicle} propose a unified multi-task learning network that can simultaneously segment the vehicles and detect their boundaries in a fully supervision manner. Pan~\textit{et al.}~\cite{IGARSS2020instance} propose to infer the mask map on the oriented bounding box (OBB) to obtain the more accurate mask predictions, especially for the densely distributed objects. To the best of our knowledge, there is still few previous work on aerial instance segmentation with box annotations.

\section{Proposed Method}

\subsection{Overview} 
To facilitate the deep level set-based instance segmentation, we suggest a unified framework to learn the potential positive samples, i.e. boxes and mask map samples, which can be jointly trained to achieve classification, box regression and mask-level prediction. Figure~\ref{fig1overallnetwork} gives the overview of the whole pipeline. Based on the backbone and feature pyramid network (FPN)~\cite{cvpr2017_FPN}, our framework encodes the deep features from the input image. We employ two head branches, including the detection branch and segmentation branch. Note that the accurate detection and instance segmentation are obtained through only the bounding box annotations. 

\textbf{Detection Branch.}
In order to better fit the spatial of aerial objects with the arbitrary orientation, the detection branch adopts the oriented bounding boxes (OBB) representation. OBB can provide the orientation information of objects for detecting the objects with arbitrary orientation and dense distribution in aerial images~\cite{cvpr2021orientedrcnn, han2021redet, iccv2019scrdet}. As the previous oriented detection methods~\cite{cvpr2019_RoI-Transformer}, the RPN regresses the rotated RoIs with five parameters $(x, y, w, h, \theta)$. Inspired by DeepSnake~\cite{cvpr2020deepsnake}, we regard the rotated RoIs as the initial contour explicitly and $K$ vertices ${\{ ({x_k},{y_k})\} _{k = 1, \ldots K}}$ are uniformly sampled along each rotated RoI. The separate $snake$ modules~\cite{cvpr2020deepsnake} are used to process the $K$ vertices to iteratively refine the contour vertices towards the ground-truth rotated bounding box. The iterative regression of contour vertices can be formulated as follows:
\begin{equation}
\{ (\Delta {x_k},\Delta {y_k})\} _{k = 1}^K = snake(\{ F({x_k},{y_k})\} _{k = 1}^K)
\end{equation}
where  $F({x_k},{y_k})$ is the concatenation
of all contour vertices' features that are obtained by the bilinear interpolation. $(\Delta {x_k},\Delta {y_k})$ are the regressed relative coordinates. With the learned offsets, we can obtain the refined vertices locations as the boundary of predicted rotated bounding box. The loss function $\mathcal{L}_{reg}$ for regression is made of Chamfer distance~\cite{cvpr2017ChamferLoss} between the regressed contour vertices and oriented ground-truth box. The classification loss function $\mathcal{L}_{cls}$ is the standard cross entropy loss based on the contour vertices' features.

\textbf{Segmentation Branch.}
The segmentation branch takes the highest-resolution feature map (i.e. $P_2$) from the feature pyramid as input and predicts the mask's probability. In order to integrate the different-resolution features, we perform the upsampling operation with the high-level semantic features (i.e. $P3$, $P4$, $P5$) to keep the same size with $P2$, and then sum them in a pixel-wise manner. Moreover, we adopt the encoder of DeepLabV3+~\cite{eccv2018_deeplabv3} to enhance the semantic features. The output feature map possesses both rich semantics and multi-spatial information for aerial images. Based on the output feature map, $N\times 1 \times \frac{H}{s} \times \frac{W}{s}$ instance feature maps $M$ are generated. $N$ is the total number of potential instances, which is same as the number of the RoIs from the RPN in the detection branch. $H \times W$ is the input size, and $s$ denotes the feature map output stride. 

Unlike Mask R-CNN using only the masks within RoIs, our proposed instance segmentation branch can generate
the full-image instance mask maps. Upon each mask map, we only regard an object as the foreground. Otherwise, it is set to background. To select the potential mask map $P_{K\times 1 \times \frac{H}{s} \times \frac{W}{s}}$ and assign the corresponding ground-truth bounding box for instance segmentation with different location, we perform the positive target assignment following the detection branch. It assigns the high-quality bounding box with ground-truth box by calculating the overlaps. This unifies the positive sample assignment measure, which is able to encode the high-quality instance-level information (e.g. the coarse shape and pose of objects) with different locations. For segmentation branch, the loss $\mathcal{L}_{seg}$ consists of two aspects, i.e., the value of level set function $\mathcal{L}_{levelset}$ with the boundary evolution and the constraint vaules $\mathcal{L}_{cons.}$ for the efficient convergence. Therefore, the instance segmentation loss can be derived as below 
\begin{equation}
\mathcal{L}_{seg} = \mathcal{L}_{levelset} + \mathcal{L}_{cons.} 
\end{equation}

\subsection{Deep Level Set for Instance Segmentation}\label{seg_loss_term}
\textbf{Level Set Formulation.}
%\subsubsection{Preliminary of Level Set}
The level set method~\cite{osher1988fronts, tip2001_active_contour} implicitly represents a parametric curve $C$, which is defined as the boundary of an open region $z$ in $\Omega$ space. The parametric curve can use the zero crossing of a levelset function $\phi (x,y):   \Omega  \to \mathbb{R}$ as 
%\begin{comment}
%\begin{equation}
%\left\{\begin{array}{l} C {\rm{ = }}\{ (x,y) \in \Omega \left| {\phi (x,y) = 0} \right.\}, \vspace{0.8ex}
%\\inside(C)  {\rm{ = }}\{ (x,y) \in \Omega \left| {\phi (x,y) > 0} \right.\}, \vspace{0.8ex}
%\\outside(C)   = \{ (x,y) \in \Omega \left| {\phi (x,y) < 0} %\right.\}.
%\end{array} \right.
%\end{equation}
%\end{comment}
\begin{equation}
C = \{ (x,y) \in \Omega |\phi (x,y) = 0 \}
\end{equation}
$inside(C) = \{ (x,y) \in \Omega |\phi (x,y)> 0\} $ and $outside(C)   = \{ (x,y) \in \Omega | {\phi (x,y) < 0}\}$ represent the region $z$ and the region outside $z$, respectively. The fitting energy terms ${E_1}(C)$ and ${E_2}(C)$  for the above two regions can be defined as follow,
\begin{equation}
\begin{array}{l}
{E_1}(C) = {\int_{\phi  > 0} {\left| {{u_0}(x,y) - {a_1}} \right|} ^2}dxdy  \vspace{1ex}  \\
{E_2}(C) = {\int_{\phi  < 0} {\left| {{u_0}(x,y) - {a_2}} \right|} ^2}dxdy
\end{array}
\end{equation}
where $u_0$ is a given image, and $a_1$ and $a_2$ are the averages of $u_0$ inside $C$ and respectively outside $C$, given by 
\begin{equation} 
\left\{ \begin{array}{l}
{a_1}(\phi) = average({u_0}) \ \ \ in\ \ \{ \phi  \ge 0\}  \vspace{1ex} \\
{a_2}(\phi) = average({u_0}) \ \ \ in\ \  \{ \phi  < 0\} 
\end{array} \right.
\end{equation}
The minimization of the energy terms can be viewed as an curve evolution along the descent of energy function. The boundary $C_0$ of object is the minimization of the fitting term
\begin{equation}
\mathop {\inf }\limits_C \{ {E_1}(C) + {E_2}(C)\}  \approx 0 \approx {E_1}({C_0}) + {E_2}({C_0})
\end{equation}
The length and area of $C$ with the curve evolution are represented as,
\begin{equation}
\begin{array}{l}
\begin{aligned}
Length(\phi  = 0) & = \int_\Omega  {\left| {\nabla H(\phi (x,y))} \right|dxdy}  \\ &= \int_\Omega  {\delta (\phi (x,y))\left| {\nabla \phi (x,y)} \right|dxdy}
\end{aligned} 
\vspace{1ex}  \\
Area(\phi  \ge 0) = \int_\Omega  {H(\phi (x,y))dxdy} 
\end{array}
\end{equation}
where $H$ is the Heaviside Function, $\delta$ is 1-D Dirac measure function. 
%\begin{equation}
%H(z) = \left\{ \begin{array}{l}
%1,z \ge 0\\
%0,z < 0
%\end{array} \right., \ \ \ \delta (z) = \frac{d}{{dz}}H(z)
%\end{equation}
%where

\textbf{Enlarged Box Region for Deep Level Set Evolution.}
The classical level set methods are defined and optimized only based on the low-level features, such as shape, color and texture between the object and background, which limits their performance on level set segmentation. Deep network possesses the ascendant ability to encode the high-level features.
%Classical level set methods for object segmentation are given an initial level set and an image as input. The energy function is defined and optimized only based on the low-level features, such as shape, region, color and texture between the object and background. This limits the performance of level set segmentation. Deep network possesses the ascendant ability to encode high-level features, which makes it possible to achieve superior performance in complex scene. 
To this end, our proposed method exploits the curve evolution with level set to find the similarity between the deep features and the object characteristics in aerial scene. The level set method with well-designed energy function makes it possible to achieve high-quality instance segmentation using the box annotations only. 

Given an input image $u_0(x,y)$ and horizontal bounding box $\mathcal{B}$, our goal is to employ a CNN branch to predict the object's boundary curve used in level set evolution within the enlarged box region $\mathcal{B^*}$. $\mathcal{B^*}$ is the $n$ times of the bounding box $\mathcal{B}$ region, which is the relaxation of $\mathcal{B}$ and introduces more context for levelset evolution on each object. As shown in Figure~\ref{fig1overallnetwork}, we build a segmentation branch based on the CNN feature encoder. The last output of this branch is the potential mask maps with the shape of $N \times 1 \times \frac{H}{s} \times \frac{W}{s}$, where $N$ is the number of predicted instances. For each instance, we only focus on the assigned enlarged bounding box region $\mathcal{B^*}$ . So, we regard its outside as the background. To better incorporate the CNN with the level-set method, we treat the values of potential mask map within $\mathcal{B^*}$ box region as the level set $\phi$, and the pixel space of input image $u_0(x,y)$ is referred as $\Omega$. $C$ is the segmentation boundary with $\phi = 0$. To generate the precise object boundary with level-set method, we design an energy function. With the minimization of this energy function, the network learns a level set $\phi$ to achieve the boundary evolution during the training phase. The energy function is defined as follows:
\begin{equation}
\begin{aligned}
L(&{C_1}, {C_2},\phi ,{\rho _{cls}}, \mathcal{B^*})  \\ & =  {\alpha _1}\sum\limits_n {{\rho _{cls}}} {\int_{\Omega \in \mathcal{B^*}}  {\left| {u_0^*(x,y) - a_1^ * } \right|} ^2}\sigma (\phi {(x,y)})dxdy \\ & +
{\alpha _2}\sum\limits_n {{\rho _{cls}}} {\int_{\Omega \in \mathcal{B^*}}  {\left| {u_0^*(x,y) - a_2^ * } \right|} ^2}(1 - \sigma {(\phi (x,y))})dxdy
\\ & +  \lambda  * Length(\phi ) + \mu Aera(\phi ) 
\end{aligned}
\label{energyfunction}
\end{equation}
where $u_0^*(x,y)$ denotes the input image after the normalization. $\sigma$ denotes $sigmoid$ function that is treated as the characteristic function for level set $\phi$.  Differently from the traditional Heaviside function $H$, $sigmoid$ function is much smoother, which is able to better express characteristics of the predicted instance. This further benefits the convergence of level set during the training process. The first two items in Eq.~(\ref{energyfunction}) force the predicted mask map to be uniform both inside and outside
object regions. $a_1^*$ and $a_2^*$ play as the average values for $inside(C)$ and $outside(C)$, respectively. Regarding to $a_1^*$ and $a_2^*$, these two terms can be expressed as below
\begin{equation}
\begin{aligned}
&{a_1^*}(\phi ) = \frac{{\int_{\Omega \in \mathcal{B^*}}  {{u_0^*}(x,y)\sigma(\phi (x,y))dxdy} }}{{\int_{\Omega \in \mathcal{B^*}} {\sigma(\phi (x,y))dxdy} }} \\
&{a_2^*}(\phi ) = \frac{{\int_{\Omega \in \mathcal{B^*}}  {{u_0^*}(x,y)(1 - \sigma(\phi (x,y)))dxdy} }}{{\int_{\Omega \in \mathcal{B^*}} {(1 - \sigma(\phi (x,y)))dxdy} }}
\end{aligned}
\end{equation}

The last two terms of Eq.~(\ref{energyfunction}) denotes the the length of the evolution boundary and inside boundary region area, respectively. They can be viewed as the regularization term to keep the boundary of segmentation map smoothing, which are denoted as follows
\begin{equation}
\begin{array}{l}
Length(\phi ) = \int_{\Omega \in \mathcal{B^*}}  {\left| {\nabla \sigma (\phi (x,y))} \right|dxdy} \vspace{1ex} \\
Area(\phi ) = \int_{\Omega \in \mathcal{B^*}}  {\sigma (\phi (x,y))dxdy} 
\end{array}
\end{equation}
It is important to note that $\rho _{cls}$ is the class-wise weighted term for the different categories, which needs various degrees of level set evolution for the different object characteristics. This parameter can be adaptive tuned online or predefined. The energy function $L$ can be easily optimized with the gradient back-propagation during training. 
The derivative of energy function $L$ upon $\phi$ can be written as below
\begin{equation}
\begin{aligned}
\frac{{\partial L}}{{\partial \phi }} & {\rm{ = }}\nabla \sigma (\phi )[{\alpha _1}{\rho _{cls}}{ ({u_0^* - a_1^*}) ^2} - {\alpha _2}{\rho _{cls}}{ ({u_0^* - a_2^*})^2} \\ &+ \lambda div(\frac{{\nabla \phi }}{{\left| {\nabla \phi } \right|}}) + \mu ]
\end{aligned}
\end{equation}

The whole evolution process is differentiable, which can be trained in an end-to-end manner. During the training process, we set the energy function $L$ as the loss function $\mathcal{L}_{levelset}$ for level set evolution. Practically, we set ${\alpha _1} = {\alpha _2} = 0.001$, $\lambda=0.00001$ and $\mu =0.000001$ to maintain the loss at a reasonable value while keeping the effective gradient propagation and balanced converge stably for each loss term. The class-wise parameter $\rho _{cls}$ is discussed in the experiment section.

\subsection{Box and Background Constraints}
With the box annotations, we can obtain the two key rules. The first one is that the predicted mask map should be limited within its ground-truth box region. The second rule is that all region outside ground truth box should be the background for the current instance. To this end, we introduce the constraints to effectively improve the convergence and prevent the interference of noise information during the training phase.

As in~\cite{cvpr2020_boxinst}, we firstly employ the coordinate projection of ground truth box to $x$-axis and $y$-axis and calculate the difference between predicted mask map and ground-truth box. Let $b^f \in {\{ 0,1\} ^{H \times W}}{\rm{ }}$ denote the binary region generated by assigning one to the locations in the ground-truth box, and zero otherwise. The mask score predictions ${m^p} \in {(0,1)^{H \times W}}$ for each instance can be regarded as the foreground probabilities. 
We define the box constraint term $\mathcal{L}_{cons.}^{box}$ as follows:
\begin{equation}
\begin{aligned}
\mathcal{L}_{cons.}^{box} & = \mathcal{L}_{proj}({m_x^p}, b_x^f) + \mathcal{L}_{proj}({m_y^p}, b_y^f)
\end{aligned}
\label{boxconstraint}
\end{equation}
\begin{comment}
\begin{equation}
\begin{aligned}
\mathcal{L}_{cons.}^{box} & = \mathcal{L}_{proj}(({{}m_x^p}), Proj{_x}(b)) \\ &+ \mathcal{L}(Pr oj{_y}({m^p}), Proj{_y}(b)),
\end{aligned}
\label{boxconstraint}
\end{equation}
\end{comment}
where ${m_x^p}$, $b_x^f$ and ${m_y^p}$, $b_y^f$ denote the $x$-axis projection and $y$-axis projection for mask prediction $m^p$ and binary ground-truth region $b^f$, respectively. The  projection can be implemented by a $\max$ operation for each axis.

The region outside the box is regarded as the background. Let  $b^b \in {\{ 0,1\} ^{H \times W}}$ denote the binary background region. We assign one to all locations outside the box region for background scores prediction. Otherwise, zero is set to the location inside the box region. The background prediction score is $m^b = \mathbbm{1} - m^p$.
%also can be represented as follows:
%\begin{equation}
%m^b = \mathbbm{1} - m^p
%\end{equation}
Therefore, we define the background constraint term $\mathcal{L}_{cons.}^{back.}$ as follows:
\begin{equation}
\mathcal{L}_{cons.}^{back.} = \mathcal{L}(m^b, b^b)
\label{backconstraint}
\end{equation}
where $\mathcal{L}( \cdot , \cdot )$ in the Eq.~(\ref{boxconstraint}) and Eq.~(\ref{backconstraint}) denotes the pixel-wise dice loss~\cite{ic3dv2016_dice_loss}. Therefore, the constraint loss $\mathcal{L}_{cons.}$ is defined as below:
\begin{equation}
{\mathcal{L}_{cons.}} = \mathcal{L}_{cons.}^{box} + \mathcal{L}_{cons.}^{back.}
\end{equation}

\section{Experiments}
\subsection{Datasets}
\textbf{iSAID.} iSAID dataset~\cite{cvprw2019isaid} consists of 2,806 HSR remote sensing images, which are collected from multiple aerial platforms. The original image sizes range from 800 $\times$ 800 pixels to 13000 $\times$ 4000 pixels. The iSAID dataset provides 655,451 instances annotations,  which is the largest dataset for instance segmentation in the HSR remote sensing imagery. 
It consists of common 15 categories: Ship, Storage Tank (ST), Baseball Diamond (BD), Tennis Court (TC), Basketball Court (BC), Ground Track Field (GTF), Bridge, Large Vehicle (LV),  Small Vehicle (SV), Helicopter (HC), Swimming Pool (SP), Roundabout (RA), Soccer Ball Field (SBF), Plane, and Harbor. The dataset are made of 1,411 training images, 458 validation images, and 937 testing images.  We make use of the predefined training set to train our models and evaluate them on the validation set. This is because the labels of testing set are unavailable.

\textbf{Potsdam.} Potsdam dataset$\footnote{https://www2.isprs.org/commissions/comm2/wg4/benchmark/2d-semlabel-potsdam/}$ is the high-resolution remote sensing dataset with the semantic labels, which consists of 38 high-resolution aerial images of 5cm GSD with 6000 $\times$ 6000 pixels. We randomly partition all the images into 28 images as the training set and 10 images as the testing set with the approximate ratio of 3:1. This dataset consists of six semantic categories, we only use the category of Car to evaluate the performance of instance segmentation, instead of using other stuffs related to scene understanding.

\subsection{Implementation Details}
We adopt ResNet-50~\cite{ResNet-cvpr2016} by default. For data preprocessing, we crop the original images into $800\times 800$ patches using a sliding window striding of 600 pixels. For fair comparison, we employ the mask Average Precision (AP)~\cite{ijcv2010_voc} as the main metric for instance segmentation to evaluate the proposed method if unspecified. For the setting of RPN, we follow RoI-Transformer~\cite{cvpr2019_RoI-Transformer}. For the training, we adopt the same training schedules
as $mmdetection$~\cite{arxiv2019mmdetection}. The optimizer used for training is SGD. The initial learning rate is set to 0.005, which is divided by 10 at each decay step. The momentum and weight decay are 0.9 and 0.0001, respectively. We train the models in 1$\times$ with 12 epochs or  2$\times$ with 24 epochs. In terms of the data augmentation, random flip and rotation are adopted during training. We conducted experiments on 4 RTX 2080Ti GPUs with a total batch size of 8 for training and used a single RTX 2080Ti GPU for inference.

For levelset evolution in the instance segmentation branch,  we employ the horizontal bounding box (HBB) annotations. Meanwhile, we use the annotations of oriented bounding box (OBB) for the detection branch. More importantly, any mask-level annotations are not used to train our models.

\subsection{Ablation Study}
Firstly, we conduct a series of ablation experiments on Postdam dataset to evaluate the effectiveness of our proposed method. Note that we train all the ablation experiments with 1$\times$ schedule, where R-50-FPN is the backbone.

\begin{figure}[t]
\centering
\includegraphics[width=1.0\linewidth]{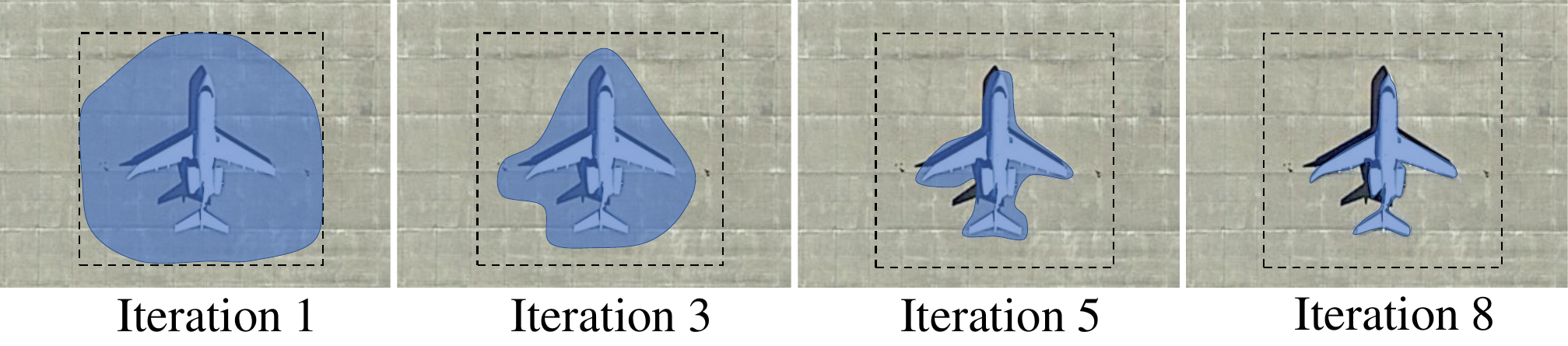}
\caption{The curve evolution process of levelset at different iteration.}
\label{levelset_evolution}
\end{figure}

\textbf{Level set and constraint loss for instance segmentation.}
We evaluate the effectiveness of level set and constraint loss under different settings. Table~\ref{tab:levelsetablation1} shows the experimental results.  It achieves only 17.8\% AP performance with the constraint loss alone, which indicates that the box and background constraint are effective. With the level set loss term, it gains the 25.3\% performance with 7.5\% improvement compared with the result using constraint loss, which shows that the proposed region-based level set method can be learned in weakly supervision using box annotations. Incorporating constraint loss with level set loss term achieves the best performance with 34.8\% AP, 57.1\% AP$_{50}$ and 40.9\% AP$_{75}$. The experimental results demonstrate that the combination of these two loss terms are effective for the aerial instance segmentation in the box-supervision manner.
\begin{table}[htbp]
	\centering
	\small
	%\footnotesize
	%\scriptsize
	%\tiny
	\setlength{\tabcolsep}{4.0 mm}{
		\begin{tabular}{c|c|ccc}
			\toprule
			$\mathcal{L}_{const.}$ & $\mathcal{L}_{levelset}$ & AP & AP$_{50}$ & AP$_{75}$ \\
			\midrule
			%\multicolumn{2}{c|}{horizontal box mask}  & \\ 
			%\multicolumn{2}{c|}{rotated box mask}  & \\ \hline
			\checkmark &   & 17.8 & 51.8 & 5.9  \\
			& \checkmark & 25.3 & 56.0 & 21.3 \\
			\checkmark & \checkmark &  \textbf{34.8} & \textbf{57.1} & \textbf{40.9} \\          
			\toprule
	\end{tabular}}
	\caption{Performance analysis of level set and constraint loss on Potsdam dataset.}
	\label{tab:levelsetablation1}
\end{table}
\begin{table}[htbp]
	\centering
	\small
	%\footnotesize
	%\scriptsize
	%\tiny
	\setlength{\tabcolsep}{4.0mm}{
		\begin{tabular}{c|ccc}
			\toprule
			Enlarged box region $\mathcal{B^*}$ & AP & AP$_{50}$ & AP$_{75}$ \\
			\midrule
			1.0$\times$& 33.2 & 59.8 & 36.0  \\
			1.5$\times$& 34.4 & 56.7 & 40.3 \\
			2.0$\times$& \textbf{36.1} & \textbf{60.4} & \textbf{42.1} \\
			2.5$\times$& 34.8 & 57.1 & 40.9\\
			\toprule
	\end{tabular}}
	\caption{Comparisons on Potsdam dataset with the different enlarged regions.}
	\label{tab:ablation2_enlaredregion}
\end{table}

\begin{table*}[t]
	\centering
	\tabcolsep=2pt
	%\resizebox{\textwidth}{20mm}{
	%% \tablesize{} %% You can specify the fontsize here, e.g., \tablesize{\footnotesize}. If commented out \small will be used.
	%\scriptsize
	\footnotesize
	\setlength{\tabcolsep}{0.75mm}{
		\begin{tabular}{c|cc|ccccccccccccccc|ccc}
			\toprule
			%Methods & backbone & PL& BD& BR& GTF& SV& LV& SH& TC& BC& ST& SBF& RA& HA& SP& HC & AP & AP$_{50}$ \\
			Methods &backbone&sched. &Ship &ST & BD& TC & BC& GTF & Bridge & LV& SV& HC& SP& RA&SBF&Plane&Harbor& AP & AP$_{50}$ & AP$_{75}$  \\
			\midrule
			\textit{fully supervised methods}: & \\
			Mask R-CNN~\cite{iccv2017maskrcnn} & R-50-C4 & $1 \times$ & 32.5 & 30.0 & 45.3 & 73.2 & 30.0 & 23.1 & 14.8 & 30.2 & 10.2 & 5.4 & 25.2 & 26.6 & 37.4 & 30.3 & 17.0 & 28.8 & 51.8 & 27.7  \\
			%Mask R-CNN~\cite{tpami2020_maskrcnn} & R-50-FPN &  \\
			%PANet~\cite{} & R-50-FPN&  \\
			PolarMask~\cite{cvpr_2020polarmask} & R-50-FPN &$1 \times$& 35.7 &32.5 &44.4 & 74.4 & 37.5 & 13.4 & 10.8 & 30.0 & 8.5 & 3.4 & 24.5 & 29.6 & 32.3 & 21.6 & 9.6 & 27.2 & 48.5 & 27.3  \\ 
			CondInst~\cite{eccv202_condinst} & R-50-FPN & $1 \times$ & 35.0 & 32.0 & 44.5 & 72.4 & 28.6 & 17.6 & 14.5 &33.7 & 10.5 & 4.6 & 26.4 & 27.9 & 34.4 & 35.7 & 24.1 & 29.5 & 54.2 & 28.3   \\
			%YOLACT~\cite{} & \\
			%SOLOv2~\cite{} &R-50-FPN & $1 \times$\\
			%$RefineMask~\cite{cvpr2021refinemask} & %R-50-FPN & $1\times$ & 42.5 & 42.0 & 55.5 & 79.1 & 41.8 &29.6 & 21.3&36.3  & 15.6 &6.2 &31.8 & 34.5 & 53.5& 46.6& 29.3& 37.7 & 59.8 &40.7  \\ 
			\hline
			\textit{box-supervised methods}: & \\
			BBTP~\cite{nips2019-bbtp} &R-50-FPN  &$1 \times$ & 17.6 & \textbf{33.9} & \textbf{42.5} & 49.6 & 26.5 & 18.3 & 6.0 & 8.7 & \textbf{11.0} & 1.3 & 19.1 & \textbf{29.8} & 28.7 & 0.8 &2.3 & 19.7 & 40.8 & 17.5\\	
			BoxInst~\cite{cvpr2020_boxinst} &R-50-FPN &$1 \times$ & 21.6 & 21.5 & 35.9 &42.5 &19.4 &15.2 & 3.5 & 19.3 & 5.6 &\textbf{1.4} & \textbf{19.3} &13.0 &19.1 & \textbf{20.5} & 9.3& 17.8 & 41.4 & 12.9 \\
			%BoxInst~\cite{cvpr2020_boxinst} &R-50-FPN &$1 \times$ & 21.8 & 23.3 & 34.8 &45.4 &17.4 &11.2 & 3.1 & 19.7 & 5.1 &\textbf{3.0} & 18.0 &10.9 &15.8 & \textbf{20.0} & 9.3& 17.2 & 41.7 & 11.7 \\
			
			Ours & R-50-FPN &$1 \times$&\textbf{24.6} & 24.4 & 37.2& \textbf{62.8} & \textbf{35.5} &\textbf{18.9} & \textbf{14.0} & \textbf{28.9} & 9.7 & 1.3 & 17.8 & 21.8 & \textbf{28.9} & 6.0 & \textbf{15.2} & \textbf{23.1} & \textbf{48.4} & \textbf{19.4} \\
			%\textbf{Ours} & R-101-FPN & $3 \times$ \\
			\toprule
	\end{tabular}}
	\caption{Class-wise instance segmentation results on iSAID \textit{val} set. }
	\label{tab:iSAIDdataset}
\end{table*}

\begin{figure*}[htbp]
	\centering
	\includegraphics[width=1.0\linewidth]{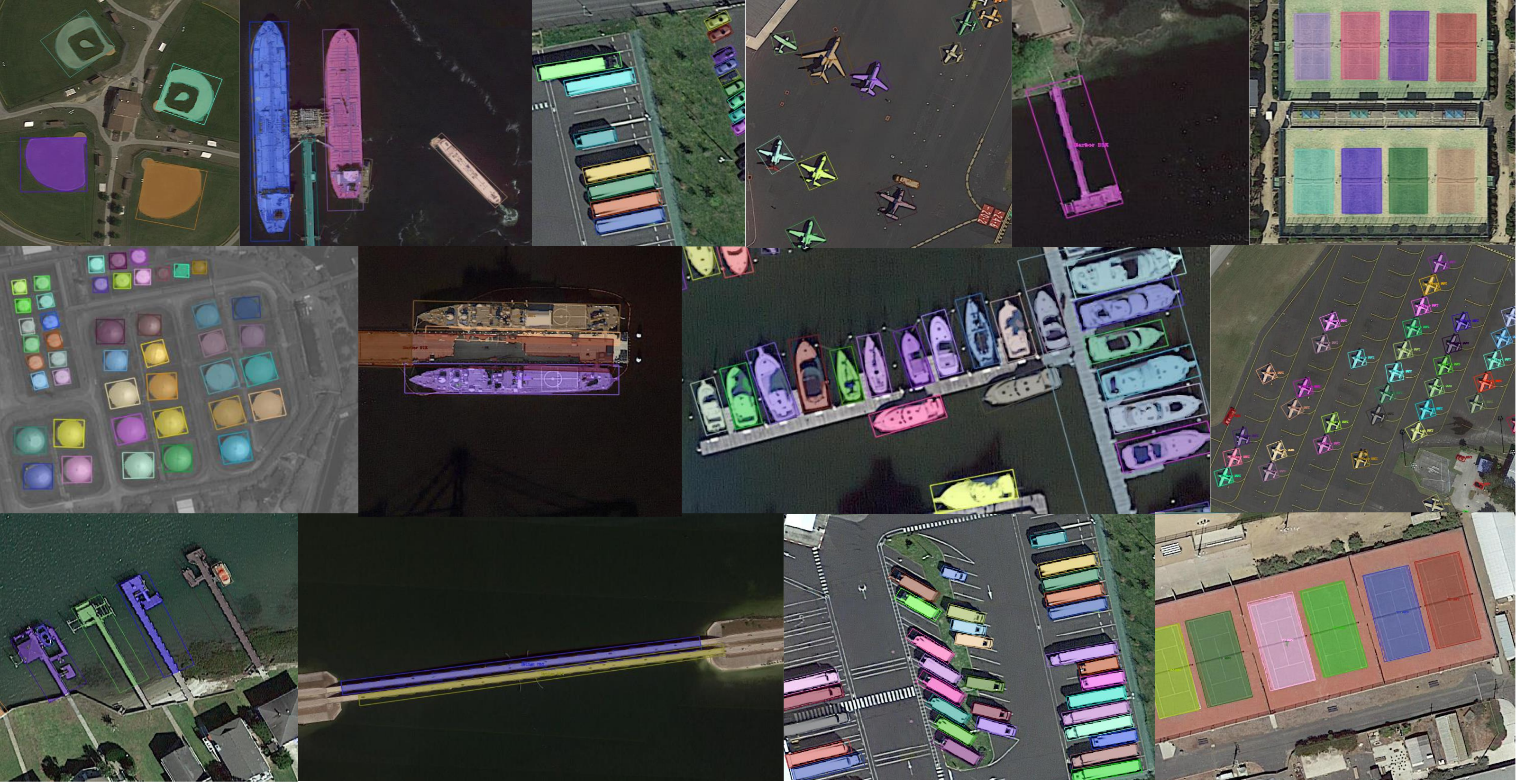}
	\caption{Some qualitative results of the proposed method on iSAID \textit{test} set.}\label{vis_results}
\end{figure*}

\textbf{Enlarged box region for level set evolution.} As shown in Table~\ref{tab:ablation2_enlaredregion}, we measure the performance of different enlarged regions for level set evolution.  The model can achieve 33.2\% AP with the 1.0$\times$ region, which is original box area. The best performance with 36.1\% AP, 60.4\% AP$_{50}$ and 42.1\% AP$_{75}$ is achieved with 2.0$\times$ region. It indicates that the larger enlarged region is beneficial for the level set evolution, which introduces more characteristic information to be learned. However, the performance begin to slightly drop when the enlarged region is set to 2.5$\times$, which may induce the noisy information with a large area. 

\textbf{Effectiveness of class-wise parameters for level set evolution.}
We further study the effects of class-wise $\rho _{cls}$ in Eq.~(\ref{energyfunction}) with different settings. In Table~\ref{tab:ablation3_calsswise}, we set different $\rho _{cls}$ for curve evolution on the Potsdam dataset, which indicates that this class-wise parameter can influence the performance of level set. When $\rho _{cls}= 0.65$, the mask AP has the better performance with 34.8\% AP. We also set a single fully connection layer on the potential mask map to adaptively learn $\rho _{cls}$ during the training phase in order to avoid the manually adjusted parameters. The results show that the self-adaptive $\rho _{cls}$ scheme can achieve the best performance with 36.5\% AP.
\begin{table}[htbp]
	\centering
	\small
	%\footnotesize
	%\scriptsize
	%\tiny
	\setlength{\tabcolsep}{3.0 mm}{
		\begin{tabular}{c|cccccc}
			\toprule
			$\rho _{cls} $ & 1.50 & 1.0 & 0.65 & 0.35 & self-adaptive \\
			%			$\rho _{cls} $ & AP & AP$_{50}$ & AP$_{75}$ \\
			\midrule
			AP & 30.8 & 31.1  &   34.8 & 33.2 & \textbf{36.5}  \\
			AP$_{50}$ & 57.4 & 56.2   & 57.1  & 59.8 & \textbf{63.2} \\
			AP$_{75}$ &  31.7 & 32.5 & \textbf{40.9} &36.0 & 40.6\\
			\toprule
	\end{tabular}}
	\caption{Analysis of the class-wise parameters for level set evolution on Potsdam dataset.}
	\label{tab:ablation3_calsswise}
\end{table}
\begin{table}[htbp]
	\centering
	\footnotesize
	%\scriptsize
	%\tiny
	\setlength{\tabcolsep}{2.0 mm}{
		\begin{tabular}{c|cccc}
			\toprule
			Samples assignment methods & AP & AP$_{50}$ & AP$_{75}$ & AP$_{box}$  \\
			\midrule
			Max-IoU~\cite{nips2015fasterrcnn} & 34.8 & 57.1 & 40.9 & 35.4\\
			ATSS~\cite{cvpr2020atss} & \textbf{36.1} & \textbf{60.4} & \textbf{42.1} & \textbf{37.3}  \\
			\toprule
	\end{tabular}}
	\caption{Effects of different samples assignment methods.}
	\label{tab:ablation4target assignment}
\end{table}

\begin{figure*}[t]
	\centering
	\includegraphics[width=1.0\linewidth]{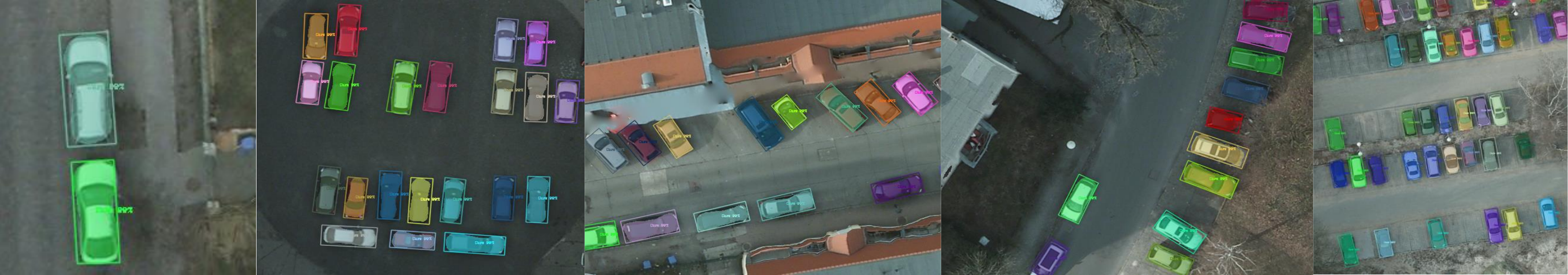}
	\caption{Visualization results of the proposed method on Potsdam dataset.}\label{potsdam_results}
\end{figure*}
\textbf{Comparisons with different samples assignment methods.} In our framework, the segmentation module follows the detection branch in the positive samples assignment. Therefore, the samples assignment method will affect the performance.
As shown in Table~\ref{tab:ablation4target assignment}, we adopt Max-IoU~\cite{nips2015fasterrcnn} and ATSS~\cite{cvpr2020atss}. A better performance with 36.1\% mask AP and 37.3\% box AP are achieved with ATSS.  Besides Table~\ref{tab:ablation4target assignment}, we employ the same Max-IoU method as other approaches without ATSS to facilitate the fair comparisons.

\subsection{Comparisons with the State-of-the-art Methods}
\textbf{Results on iSAID.} iSAID is the de-facto benchmark dataset for aerial instance segmentation. We compare our proposed approach against the state-of-the-art methods with the different forms of supervision. Table~\ref{tab:iSAIDdataset} lists the experimental results. It can be clearly observed that our approach outperforms the state-of-the-art box-supervision method BoxInst~\cite{cvpr2020_boxinst} for general scene by 5.3\% AP, 7.0\% AP$_{50}$ and 6.5\% AP$_{75}$, respectively. In addition, our method performs better than BBTP~\cite{nips2019-bbtp} over 3.4\% AP, 7.6\% AP$_{50}$ and 1.9\%  AP$_{75}$.
%Especially for some categories, such as Tennis Court (TC), Basketball Court (BC), Bridge, Roundabout (RA) and Soccer Ball Field (SBF), the proposed method gains over improvements 10\% AP.
Furthermore, our method demonstrates the promising performance comparing to the top-performing fully-supervised instance segmentation methods using mask-level annotations. For the categories of Basketball Court (BC), Ground Track Field (GTF), Bridge and Large Vehicle (LV), the proposed approach achieves the competitive results. These encouraging results show that the proposed deep level set-based method is effective for the task of aerial instance segmentation by curve evolution, which greatly narrows down the performance gap between the fully mask-supervised and weakly box-supervised instance segmentation.  Qualitative results on iSAID are shown in Figure~\ref{vis_results}.

\textbf{Results on Potsdam.}
As shown in Table~\ref{tab:PotsdamResults}, we evaluate the recent segmentation methods with 2$\times $ training schedules on the Potsdam dataset, which contains the urban scene with the challenging car segmentation. It can be clearly seen that our presented method achieves the best performance of 47.3\% AP, which outperforms the box-supervised methods like BBTP~\cite{nips2019-bbtp} and BoxInst~\cite{cvpr2020_boxinst} over 17.6\% AP and 8.9\% AP, respectively. Comparing to the mask-supervised methods, the proposed method achieves the comparable results with narrow gap.
Figure~\ref{potsdam_results} shows some visualization results on Potsdam dataset.   

\begin{table}[htbp]
	\centering
	\small
	%\footnotesize
	%\scriptsize
	%\tiny
	\setlength{\tabcolsep}{1.0mm}{
		\begin{tabular}{c|cc|ccc}
			\toprule
			Methods & backbone & sched. & AP & AP$_{50}$ & AP$_{75}$ \\
			\midrule
			\textit{fully supervised methods}: & \\ [2pt]
			Mask R-CNN~\cite{iccv2017maskrcnn} & R-50-C4 & $2 \times$ & 51.7 & 72.6 & 62.1   \\ [1.5pt]
			
			PolarMask~\cite{cvpr_2020polarmask} & R-50-FPN &  $2\times$ & 50.0 & 71.0 & 59.9 \\[1pt]
			CondInst~\cite{eccv202_condinst} & R-50-FPN & $2\times$ & 53.6 & 74.2 & 64.1 \\ [1pt]
			
			\hline
			\textit{box-supervised methods}: & \\ [2pt]
			BBTP~\cite{nips2019-bbtp}  & R-50-FPN & $2 \times$ & 29.7& 69.2 &19.5  \\ [1.5pt]
			BoxInst~\cite{cvpr2020_boxinst} & R-50-FPN & $2\times$ & 38.4 & 71.4 & 39.2  \\ [1.5pt]
			Ours & R-50-FPN & $2 \times$  &\textbf{47.3} & \textbf{71.9} & \textbf{44.7} \\ [1.0pt]
			\toprule
	\end{tabular}}
	\caption{Comparisons with state-of-the-art methods on Potsdam dataset.}
	\label{tab:PotsdamResults}
\end{table}

\section{Conclusion}
This paper has proposed an effective deep level set-based network with box supervision for aerial instance segmentation. The detection branch and segmentation branch are introduced, which are jointly training with a unified potential samples assignment manner. The well-designed energy function can evolve object accurate boundaries with complex background and prevent inter- and intra-class interference efficiently in high-resolution aerial images. Experiments results on two high-resolution aerial dataset have demonstrated that our proposed approach achieves the promising results with box supervision, and gain the comparable performance with mask-supervised methods on some categories with the large-scale aerial benchmarks. This work can promisingly narrows the performance gap between the fully supervised and box-supervised instance segmentation in aerial scene.  

For the future work, we will further extend this method to the natural scene like COCO dataset to verify its generalization and provide the analysis of its advantages and limitations.

%%%%%%%%% REFERENCES
{\small
\bibliographystyle{ieee_fullname}
\bibliography{egbib}
}

\end{document}